\documentclass[review]{elsarticle}
\usepackage{lineno, hyperref}
\modulolinenumbers[5]
\usepackage{geometry}
\journal{Journal of \LaTeX\ Templates}
\usepackage{verbatim}
\usepackage{amsfonts}
\usepackage{graphics}
\usepackage{subfigure}
\usepackage{float}
\usepackage{tikz}
\usepackage{booktabs}
\usepackage{threeparttable}
\usepackage{makecell}
\usepackage{multirow}
\usepackage{multicol}
\usepackage{appendix}
\usepackage{color}
\usepackage{amsmath}
\usepackage{verbatim}
\usepackage{algorithm}
\usepackage{algorithmic}

\biboptions{authoryear}

\begin{document}
    \begin{frontmatter}
        \title{Wetland mapping from sparse annotations with satellite image time
        series and temporal-aware segment anything model}


        \author[HKU,PCL]{Shuai Yuan}
        \author[PCL]{Tianwu Lin}
        \author[HKU]{Shuang Chen}
        \author[PCL]{Yu Xia}
        \author[PCL]{Peng Qin}
        \author[PCL]{Xiangyu Liu}
        \author[PCL]{Xiaoqing Xu}
        \author[SZU]{Nan Xu}
        \author[HKU]{Hongsheng Zhang}
        \author[PCL]{Jie Wang\fnref{myfootnote1}}
        \author[HKU]{Peng Gong\fnref{myfootnote1}}



        \address[HKU]{Department of Geography, The University of Hong Kong, Hong
        Kong, China}
        \address[PCL]{Pengcheng Laboratory, Shenzhen, China.}
        \address[HGS]{Department of Electronics and Information Engineering, Harbin
        Institute of Technology (Shenzhen), Shenzhen, China}
        \address[SZU]{Key Laboratory for Geo-Environmental Monitoring of Great Bay Area, Ministry of Natural Resources, Guangdong Key Laboratory of Urban Informatics, Shenzhen Key Laboratory of Spatial Smart Sensing and Services, Shenzhen University, Shenzhen, China}
        \fntext[myfootnote1]{Corresponding author: penggong@hku.hk, wangj10@pcl.ac.cn}

        \begin{abstract}
            \label{abstract} Accurate wetland mapping is critical for ecosystem monitoring
            and management, yet acquiring dense pixel-level annotations is
            prohibitively costly. In practice, only sparse point labels are
            typically available, and existing deep learning-based models struggle
            under such weak supervision. Meanwhile, wetlands exhibit strong seasonal
            and inter-annual dynamics, making single-date imagery insufficient
            and causing substantial omission and commission errors when mapping.
            Although powerful foundation models like the Segment Anything Model (SAM)
            provide promising generalization from point prompts, it is
            intrinsically designed for static natural images, resulting in spatially
            fragmented masks in heterogeneous wetland environments and cannot exploit
            satellite image time series. To address these challenges, we propose
            WetSAM, a novel SAM-based framework that effectively leverages
            satellite image time series to enhance wetland mapping from sparse
            point annotations. WetSAM adopts a dual-branch design: (1) The
            temporal branch is prompted by sparse point labels to extend the SAM
            with a hierarchical adapter and a dynamic temporal aggregation
            module. By decomposing time series into seasonal trends and transient
            events, this branch effectively distinguishes wetland features from
            phenological variations; (2) The spatial branch reconstructs distinct
            boundaries via a temporal-constrained region-growing strategy,
            iteratively expanding sparse points into reliable dense pseudo-labels;
            (3) A bidirectional consistency regularization enforces minimizing the
            discrepancy of the predictions from two segmentation heads of two branches.
            We validate the effectiveness of WetSAM across eight diverse global
            locations, each covering an area of around $5,000 \ km^{2}$ and with
            various wetland types and geographical features. WetSAM reaches an
            average F1-score of 85.58\%, considerably outperforming other state-of-the-art
            algorithms. Results demonstrate that WetSAM
            achieves accurate, structurally consistent segmentation from sparse
            labels. With minimal labeling effort, our framework shows strong
            generalization ability and holds promise for scalable, low-cost wetland
            mapping at high spatial resolutions.
        \end{abstract}

        \begin{keyword}
            wetland mapping, satellite image time series, sparse annotation,
            segment anything model, temporal adaptation
        \end{keyword}
    \end{frontmatter}


    \section{Introduction}
    \label{intro} Wetlands, as one of the world's most productive ecosystems, provide
    invaluable ecosystem services \citep{gong2010china, yuan2025comprehensive},
    including water balance regulation, carbon storage, and mitigation of global
    climate change \citep{zhang2021development, erwin2009wetlands, pattison2018wetlands}. 
    However, accelerating climate change and human disturbances have led to widespread
    wetland degradation \citep{hu2017global}, creating an urgent need for accurate,
    efficient, and frequently updated wetland maps to support conservation and management
    \citep{yuan2025comprehensive,yuan2025gwd30}.

    In this context, satellite remote sensing, with its unique advantages of
    synoptic coverage, multi-temporal capabilities, and periodic observations, has
    become an indispensable tool for large-scale wetland monitoring \citep{gong2010china}.
    From the decades-long historical archives provided by the Landsat series to the
    high spatiotemporal resolution data streams from modern satellite constellations
    like Sentinel and Gaofen \citep{murray2019global, wang2024interannual}, it is
    now possible to characterize both intra-annual and inter-annual wetland dynamics
    driven by seasonal hydrology and vegetation phenology. Nevertheless, despite
    their potential, current wetland mapping faces two major limitations that
    impede large-scale, high-resolution applications.

    First, data-driven deep learning methods have been increasingly applied to
    wetland mapping using medium- to high-resolution satellite images \citep{hosseiny2021wetnet,jamali20223dunetgsformer}.
    Such supervised models typically rely on dense pixel-level annotations, yet
    producing such labels for wetlands is prohibitively expensive and often
    infeasible. Unlike many man-made objects with well-defined edges \citep{yuan2024relational},
    wetland boundaries are inherently gradual, blurry, and spatially heterogeneous, reflecting transitional zones between aquatic and
    terrestrial environments. As a result, manual pixel-wise annotations across
    large areas are labor-intensive and not scalable \citep{liu2025pointsam}. In
    practice, the most realistic form of annotation is sparse point labels,
    where annotators mark representative wetland and non-wetland locations
    without attempting to delineate uncertain boundaries \citep{yuan2025dynamic}.
    However, such point-level supervision provides extremely limited spatial information
    and does not directly encode object extent or boundary geometry \citep{chan2025sparse}.
    Under this setting, existing weakly supervised segmentation models are
    fundamentally constrained: while sparse points can guide the model toward the
    semantic identity of wetlands, they offer little supervision on wetland
    boundaries, often leading to fragmented predictions when applied to complex and
    heterogeneous wetland landscapes \citep{hua2021semantic, xu2022consistency}.

\begin{figure}[t]
        \centering
        \includegraphics[width=0.9\linewidth]{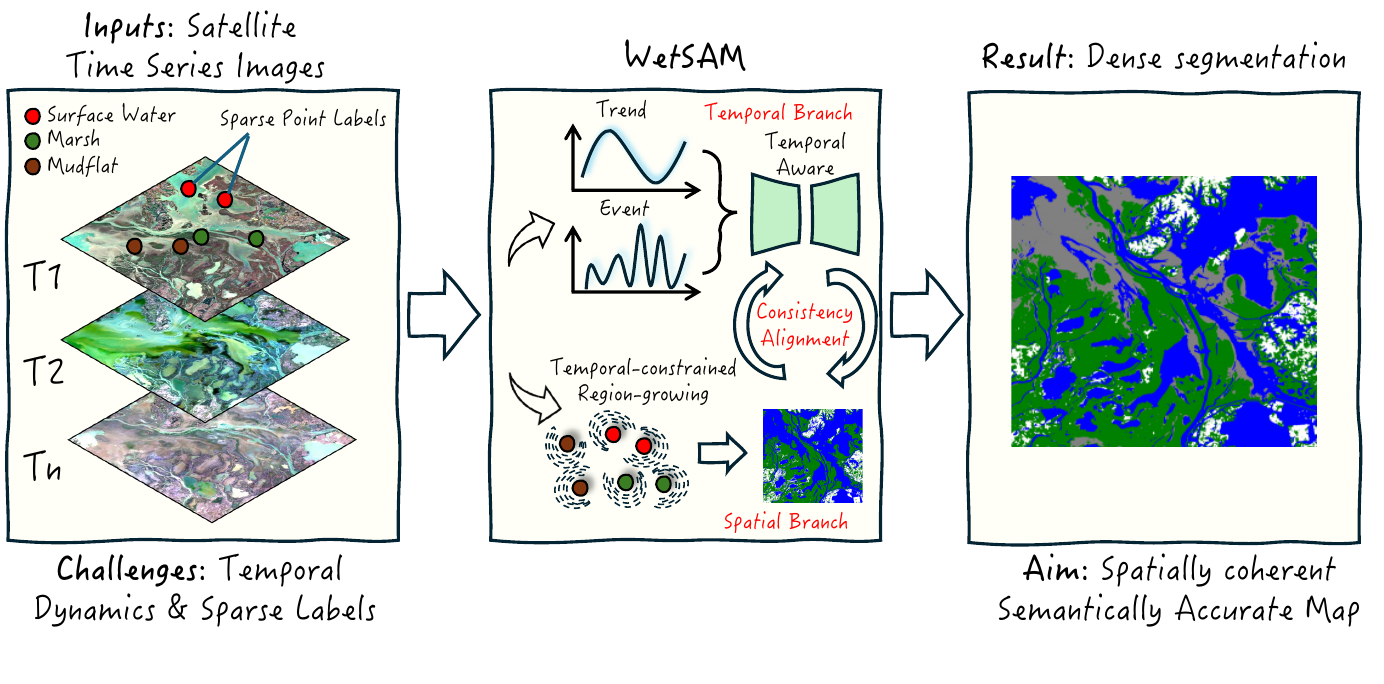}
        \vspace{-2em}
        \caption{Overview. We propose an end-to-end, single-
stage model for wetland mapping from satellite image time series under sparse point supervision. Note the difficulty of semantic segmentation of wetlands from a single image, highlighting the need for modeling temporal dynamics and spatial contexts.}
        \label{fig:intro}
    \end{figure}

    Second, wetlands exhibit strong temporal dynamics, and single-date imagery
    is insufficient for reliable classification \citep{yuan2025comprehensive}.
    Seasonal inundation, temporary flooding, and phenological changes can cause
    wetlands to appear spectrally similar to uplands during dry periods, or vice
    versa (Figure \ref{fig:intro}). A single snapshot only captures the transient
    appearance of wetlands rather than their temporal features. Incorporating time-series
    imagery is therefore essential, as it enables the full seasonal trajectory
    to reveal the characteristic dynamics of wetlands and resolve ambiguities. Although
    time-series deep learning models can exploit temporal information, they
    typically rely on full pixel-level supervision, and how to effectively incorporate
    such rich time-series information under sparse supervision remains largely unexplored
    \citep{garnot2021panoptic}. For example, most existing multi-temporal deep
    learning models for wetland mapping are trained with dense pixel-level
    annotations, while learning temporal representations from sparse point labels
    has received little attention \citep{hosseiny2021wetnet, marjani2025novel}.

    Recently, vision foundation models such as the Segment Anything Model (SAM) \citep{kirillov2023segment}
    have emerged as a promising solution to mitigate annotation costs by
    generating segmentation masks from point prompts. However, directly applying
    SAM to wetland mapping exposes two critical limitations. (1) SAM is trained on
    natural images and struggles with the complex patterns in satellite imagery,
    leading to spatially fragmented masks when prompted with sparse points
    \citep{zhang2025csw}. (2) Wetlands exhibit diverse temporal behaviors: slow-varying
    phenological trends and abrupt hydrological events. Standard SAM is
    inherently static and cannot model temporal signatures, and existing adaptations
    fail to decouple these two distinct temporal frequencies, leading to
    confusion between temporary inundation and permanent wetland features. As a result,
    SAM cannot reliably delineate wetlands nor exploit their characteristic
    seasonal trajectories. These limitations highlight a critical research gap:
    No existing framework can simultaneously learn from sparse point annotations
    and leverage satellite image time series for accurate wetland mapping, nor effectively
    adapt SAM to the temporal and spectral features of remote sensing imagery.


    To bridge this gap, we propose WetSAM, a novel end-to-end framework
    designed to effectively leverage satellite image time series to accurately
    map dynamic wetlands only under sparse point-level supervision (Figure \ref{fig:intro}). Specifically,
    WetSAM consists of three key modules. The \textbf{temporal branch} extends
    SAM by introducing a hierachical adapter and a novel dynamic temporal
    aggregation module, which decomposes satellite image time series into low-frequency
    phenological trends and high-frequency hydrological events, enabling SAM to
    capture the complex temporal characteristics of wetlands under the guidance
    of sparse point annotations and to effectively distinguish between stable
    wetland core areas and seasonal transition zones. The \textbf{spatial branch}
    addresses the limited spatial coverage of sparse supervision by transforming
    region growing from a fixed post-processing step into a supervision
    generation mechanism. Starting from sparse point annotations, a temporal-constrained
    region-growing strategy is used to progressively generate dense pseudo
    labels, which provide structured supervision for learning spatially complete
    wetland boundaries in an end-to-end manner. These pseudo-labels provide the
    model with far richer supervisory information than the original sparse
    points. Finally, a \textbf{bidirectional prediction alignment} ensures the predictions
    from the two branches are guided to be mutually reinforcing and aligned,
    ultimately enabling the model to learn to generate wetland boundaries that
    are both accurate and structurally coherent.

    The main contributions of this study are as follows:
    \begin{itemize}
        \item We propose WetSAM, the first framework that integrates sparse
            point supervision with satellite image time series for wetland
            mapping. By jointly leveraging spatial and temporal cues within a unified
            design, WetSAM alleviates the two major challenges of wetland
            mapping, i.e., extremely limited annotations and pronounced seasonal
            dynamics, without requiring dense labels.

        \item We adapt and extend the Segment Anything Model (SAM) to satellite image
            time series processing. A temporal adaptation branch extracts
            wetland phenological features from time series imagery, while a spatial
            branch generates reliable dense supervision via a temporal-constained
            region-growing strategy. A bidirectional consistency regularization
            further aligns temporal and spatial predictions, enabling accurate and
            structurally coherent segmentation from sparse labels.

        \item We demonstrate the scalability and generalization ability of WetSAM
            across eight diverse wetland regions globally. Extensive experiments
            show that WetSAM substantially outperforms existing baselines under
            minimal annotation cost, providing a practical and cost-effective solution
            for high-resolution wetland mapping at regional to global scales.
    \end{itemize}

    The rest of this paper is organized as follows. Related work of this study is
    introduced in Section \ref{sec:related_work}. Section \ref{sec:wetsam}
    illustrates WetSAM in detail. Section \ref{sec:dataset} introduces the study
    areas and materials in this work. Section \ref{sec:results} shows the
    experimental results. Ablation studies and discussions are presented in Section
    \ref{sec:ablation}. Conclusions are summarized in Section
    \ref{sec:conclusion}.

    \section{Related Work}
    \label{sec:related_work}
    \subsection{Wetland mapping using satellite remote sensing imagery}

    Wetland mapping via satellite remote sensing has evolved from traditional
    image processing to machine and deep learning. Early methods relying on handcrafted
    features and spectral indices \citep{xu2018comparison, ashok2021monitoring, dong2014mapping}
    were interpretable but sensitive to atmospheric and spectral variations. Subsequent
    approaches, including rule-based strategies \citep{dronova2015object, ruan2012mapping},
    object-based analysis \citep{wang2023wetland}, and spectral unmixing \citep{hong2021interpretable,sidike2019dpen},
    addressed some limitations but often lacked generalization and required
    manual parameterization. Machine learning algorithms, such as SVM \citep{lin2013remote}
    and random forests \citep{mahdianpari2017random}, later improved accuracy and
    flexibility by modeling nonlinear relationships \citep{jing2020exploring}. However,
    these methods remain dependent on representative training samples and are
    limited in exploiting time-series information \citep{yuan2025comprehensive}.

    Due to powerful feature representations, deep learning has since emerged as
    the dominant approach, providing remarkable capabilities for automatic feature
    extraction and large-scale mapping. Convolutional and fully convolutional
    networks \citep{pouliot2019assessment, li2022spectral}, together with region-proposal
    methods such as Mask R-CNN \citep{guirado2021mask} and Transformer-based architectures
    \citep{jamali2022swin}, have demonstrated clear advantages in handling high-resolution
    data, preserving complex boundaries, and integrating multi-source information.
    These advances have significantly raised the accuracy ceiling for wetland
    classification and segmentation. For example, \citet{ma2022identifying} proposed
    cascaded R-CNN model to detect the dike-ponds with manual regular boundaries.
    \citet{jamali2022swin} combined Swin transformer and deep CNN for complex coastal
    wetland classification.

    Yet, despite their success, two fundamental limitations remain unresolved.
    First, current deep learning frameworks rely heavily on dense, pixel-level
    annotations for supervision. Such labels are prohibitively expensive to acquire
    for wetlands, where boundaries are often fuzzy, transitional, and context-dependent.
    In practice, point-level annotations are far easier to obtain \citep{wang2023wetland, zhang2024global}, but existing models
    are ill-equipped to learn effectively from such sparse supervision. Second, wetlands
    exhibit strong seasonal and inter-annual dynamics driven by hydrology and vegetation
    phenology, therefore single-date imagery often fails to capture these
    dynamics. Although time-series deep learning models have shown the ability to
    extract temporal patterns \citep{garnot2021panoptic, marjani2025novel}, they
    rely on dense supervision and are therefore unsuitable for the sparse-label setting
    common in wetland mapping.

    These two limitations directly motivate the development of new frameworks that
    can both unlock the value of sparse point annotations and explicitly model temporal
    dynamics.

    \subsection{SAM in remote sensing}
    Recent advances in vision foundation models, particularly the Segment
    Anything Model (SAM), have created new opportunities to reduce annotation cost
    \citep{kirillov2023segment}. SAM can generate segmentation masks from simple
    prompts such as points or boxes, demonstrating impressive generalization
    across natural images. Numerous studies have been conducted to further explore
    the capabilities of SAM and to apply it effectively across diverse task scenarios
    \citep{ma2024segment, yan2023ringmo, chen2023sam, zhang2023faster}.

    However, its potential may be constrained in certain domain-specific conditions
    \citep{zhang2023comprehensive, ali2025review}. To address this limitation,
    SAM can be adapted or fine-tuned with even a single example of a new class
    or object to improve its performance. Building on this idea, multiple
    studies have fine-tuned or adapted SAM to enhance its effectiveness in
    remote sensing tasks.
    For example, \citet{zhou2024mesam} proposed Multiscale Enhanced SAM (MeSAM),
    a fine-tuning strategy tailored to remote sensing images that adapts SAM for
    semantic segmentation tasks and improves its segmentation accuracy on such
    data. \citet{wang2023samrs} leveraged SAM and existing remote sensing object
    detection datasets to construct an efficient pipeline for generating a large-scale
    remote sensing segmentation dataset SAMRS. \citet{yan2023ringmo} introduced
    RingMo-SAM, a foundation model for multimodal remote sensing image
    segmentation, capable of segmenting both optical and SAR data while also identifying
    object categories. This highlights SAM’s potential for deployment in remote
    sensing applications while reducing the need for extensive manual annotation.


    However, wetlands are highly dynamic ecosystems shaped by seasonal hydrology,
    climate variability, and human disturbance. While such dynamics are critical
    for reliable wetland delineation, SAM is fundamentally designed as a single-image
    foundation model and lacks native mechanisms to model temporal dependencies.
    As a result, single-date predictions are often insufficient to characterize
    intra-annual variability and periodic wetland transitions. Even though SAM 2
    and SAM 3 \citep{ravi2024sam, carion2025sam} support temporal inputs, they are
    primarily designed for continuous visual observations such as videos, where
    temporal coherence is dense and object identities evolve smoothly across
    frames. Satellite image time series, by contrast, are typically sparse and irregular,
    and are subject to strong seasonal, atmospheric, and sensor-induced
    variability, which challenges the core assumptions of video-based temporal modeling
    and limits their applicability to dynamic wetland mapping.

    Moreover, in practical wetland mapping, supervision is often limited to sparse point annotations. Although SAM supports point prompts for mask generation, its prompt-to-mask paradigm is primarily driven by local appearance similarity and assumes relatively coherent object structures \citep{kirillov2023segment}. Recent studies have shown that, when applied to remote sensing imagery, such point-prompted segmentation frequently produces spatially fragmented or globally inconsistent masks, especially in heterogeneous and densely structured scenes \citep{subhani2025resam}. This limitation is particularly evident in wetland environments. Wetlands exhibit strong spatial heterogeneity, gradual transitions, and blurred boundaries \citep{dronova2015mapping}. Under the sparse supervision, relying solely on point prompts lacks the constrains for the boundary, resulting in spatially fragmented masks. These observations suggest that directly applying SAM to wetland mapping remains challenging without additional mechanisms to enforce spatial consistency.

    Consequently, the central challenge addressed in this study is how to effectively leverage temporal information and sparse point annotations to map spatially coherent and semantically accurate wetland extents. To this end, we propose WetSAM, an end-to-end framework that extents SAM to integrate time-series information with iterative spatial refinement, enabling reliable wetland mapping from satellite image time series under sparse point supervision.

    \section{WetSAM} \label{sec:wetsam}
    \subsection{Problem definition}
    \label{sec:problem_definition}

    Let $\mathcal{X}= \{x_{t}\}_{t=1}^{T}$ denote a satellite image time series,
    where $x_{t}\in \mathbb{R}^{H \times W \times C}$ is the image at time $t$. Let
    $\mathcal{P}= \{(p_{i}, l_{i})\}_{i=1}^{N}$ be the set of sparse point labels,
    where $p_{i}\in \Omega = \{1, \dots, H\} \times \{1, \dots, W\}$ are the
    spatial coordinates and $l_{i}\in \{0, 1, ..., K\}$ is the corresponding
    class label for one of $K$ wetland categories. Critically, the number of
    annotated points $N$ is significantly smaller than the total number of
    pixels ($N \ll H \times W$), framing our task as a weakly-supervised learning
    problem.

    Our goal is to train a network, parameterized by $\theta$, which functions as
    a mapping $f_{\theta}: \mathbb{R}^{T \times H \times W \times C}\rightarrow \{
    0, 1, .. ., K\}^{H \times W}$. This network takes the entire image time
    series $\mathcal{X}$ as input and outputs a dense semantic segmentation map
    $\mathcal{M}= f_{\theta}(\mathcal{X})$. Since the dense ground-truth mask is
    unavailable during training, the model's parameters $\theta$ can only be
    optimized by minimizing a loss function evaluated at the sparse annotation
    locations. The objective is formally defined as:
    \begin{equation}
        \min_{\theta}\mathbb{E}_{(\mathcal{X}, \mathcal{P})}\left[ \frac{1}{N}\sum
        _{i=1}^{N}\mathcal{L}_{\text{ce}}(f_{\theta}(\mathcal{X})[p_{i}], l_{i})
        \right], \label{eq:objective_revised}
    \end{equation}
    where $f_{\theta}(\mathcal{X})[p_{i}]$ denotes the model's predicted
    probability distribution at the spatial coordinate $p_{i}$, and $\mathcal{L}_{\text{ce}}$
    is the standard cross-entropy loss.

    Solving this objective presents two fundamental challenges that directly correspond
    to the core problems identified in wetland mapping: First, the function $f_{\theta}$
    must effectively learn to aggregate information from the entire time series $\mathcal{X}$
    to make a correct prediction at a single point $p_{i}$. This requires the
    model to comprehend the complex phenological patterns and temporal features
    of different wetland types, merely from sparse supervisory signals. Second, the
    training loss is computed only on a very small subset of pixels, which means
    propagating the semantic information from these sparse points $\mathcal{P}$
    to the entire, unlabeled image to generate a spatially coherent and structurally
    complete segmentation map $\mathcal{M}$.

    Our proposed method, WetSAM, is designed to explicitly address these two challenges
    within a unified dual-branch framework, including a temporal branch, a spatial
    branch, and a consistency regularization in between as detailed in the
    following sections. The overall framework of WetSAM is presented in Figure \ref{fig:wetsam}.

    \subsection{Temporal branch: enabling SAM time-aware}
    Although SAM demonstrates strong generalization from point prompts, it is
    inherently designed for static natural images and cannot exploit the rich
    temporal cues present in satellite image time series1. As a result, SAM
    fails to capture the characteristic phenological patterns of wetlands, which
    are essential for distinguishing wetland categories from seasonal inundation
    or spectral ambiguities2. To enable SAM to interpret temporal dynamics under
    sparse supervision, we design a two-level temporal adaptation strategy that (1)
    adapts SAM to remote sensing imagery and (2) models wetland temporal contexts
    explicitly.

\begin{figure}[t]
        \centering
        \includegraphics[width=\linewidth]{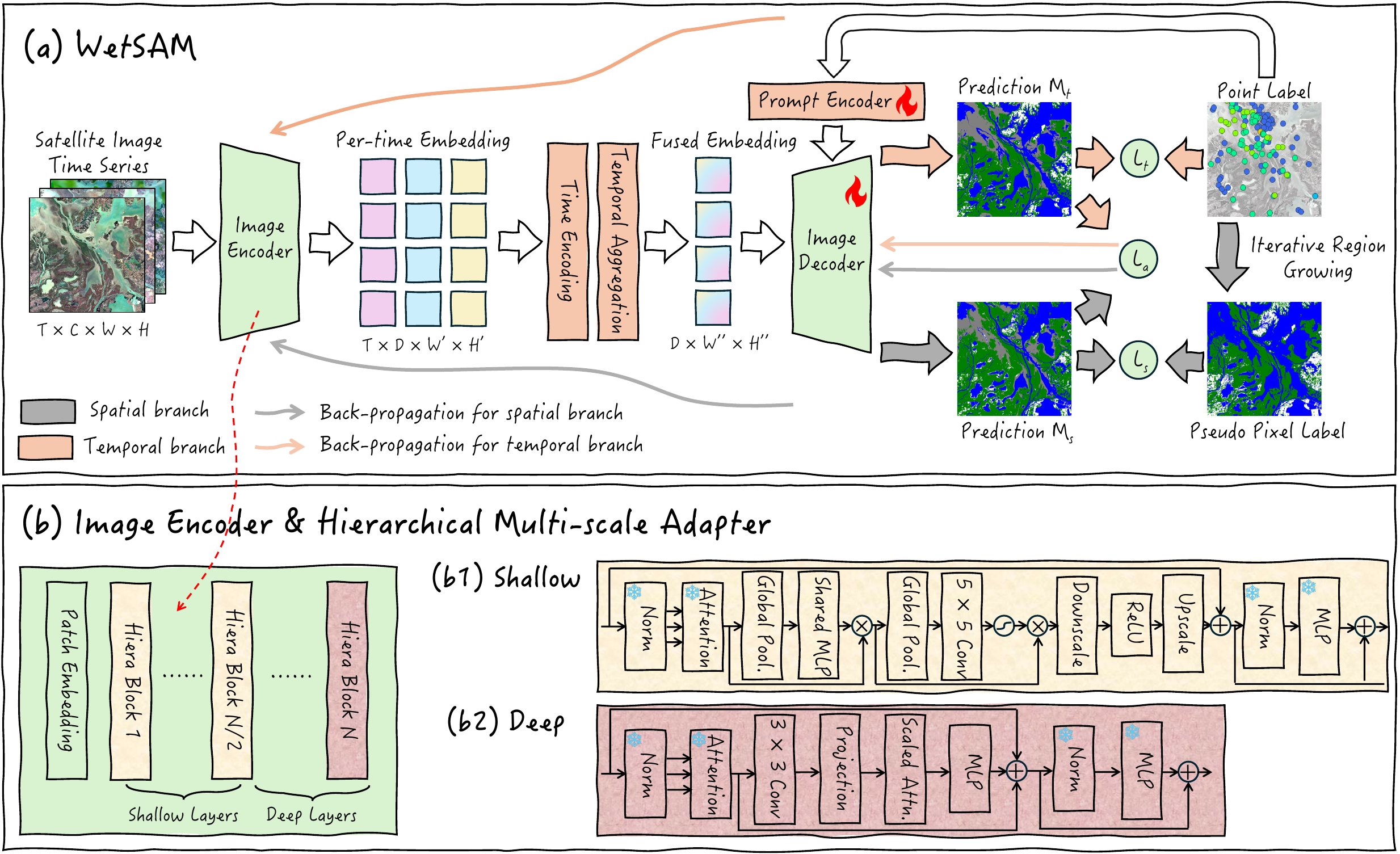}
        \vspace{-1em}
        \caption{The WetSAM framework. (a) is the overview of the framework. (b)
        is the detailed architecture of the Encoder.}
        \label{fig:wetsam}
    \end{figure}

    \textbf{Hierarchical multi-scale adapter.} Wetland remote sensing imagery exhibits unique spectral mixtures, moisture-driven reflectance changes, and
    fine-scale vegetation-water boundaries that SAM is not originally designed
    for. Meanwhile, wetland appearance varies substantially across seasons, making
    both shallow textures and deep semantics important. Existing adapter designs
    for remote sensing mainly target simple objects and are insufficient for
    wetlands, which contain high intra-class variability. To address this, we integrate
    a hierarchical multi-scale adapter into SAM’s encoder. Shallow layers focus
    on enhancing fine-scale boundary cues, while deeper layers refine temporally
    varying semantic patterns. As shown in Figure \ref{fig:wetsam} (b1) and (b2),
    for each timestamp $t$, the shallow adapter takes the feature map and
    applies (1) global pooling and two shared MLPs for spectral re-weighting, (2)
    global pooling and a $5\times5$ convolution for capturing contextual patterns,
    and (3) a bottleneck structure to reduce the dimensionality of the feature
    map. The deep adapter takes the feature map and applies a single $3\times3$ convolution
    with a scaled dot-product attention mechanism to efficient semantic refinement.
    Together, these hierarchical adapters align SAM’s representation space with remote
    sensing characteristics.

    \textbf{Dynamic temporal contexts awareness.} After extracting representations
    $\{F_{t}\}_{t=1}^{T}$ for each timestamp through the hierarchical adapter,
    we further incorporate explicit temporal positional information to model the
    phenological evolution of wetlands. Since different time points exhibit distinct
    seasonal patterns, encoding temporal position is essential for
    distinguishing wetland dynamics driven by seasonal cycles. For each timestamp
    $t$, we first introduce a sinusoidal positional embedding $E_{t}$ as follows:
    \begin{equation}
        E_{t}^{(2i)}=sin(\frac{t}{10000^{\frac{2i}{D}}}), \quad E_{t}^{(2i+1)}=co
        s(\frac{t}{10000^{\frac{2i}{D}}}),
    \end{equation}
    where $i$ indexes the embedding dimension. This formulation produces a
    continuous and periodic representation of time, enabling the model to naturally
    capture annual hydrological cycles. The time-aware feature at timestamp $t$
    is then computed as $\tilde{F}_{t}=F_{t}+E_{t}$.

\begin{figure}[t]
        \centering
        \includegraphics[width=\linewidth]{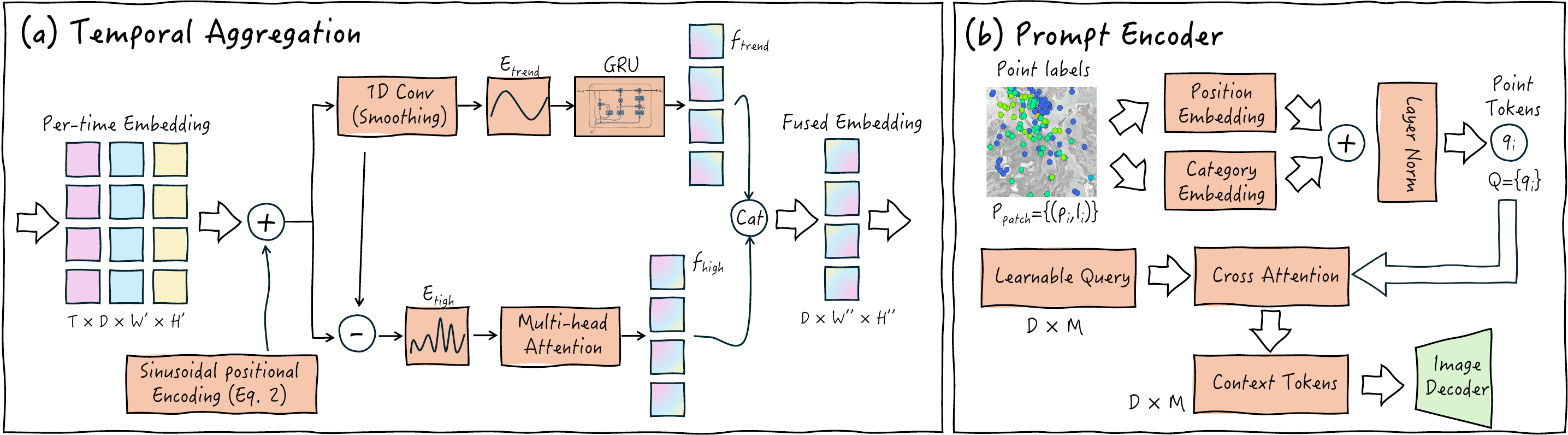}
        \vspace{-1em}
        \caption{The temporal aggregation module.}
        \label{fig:temporal}
    \end{figure}

    To effectively aggregate the temporal sequence $\{\tilde{F}_{t}\}_{t=1}^{T}$ into a
    comprehensive wetland representation, we design a temporal aggregation module
    inspired by the idea of trend–remainder decomposition in STL \citep{cleveland1990stl}, but implement
    it in a learnable and end-to-end manner. The module separates the time
    series into low-frequency seasonal trends and high-frequency transient fluctuations as shown in Figure \ref{fig:temporal} (a).
    First, we employ a learnable smoothing operation to extract the trend signal
    $E_{trend}$. This is implemented via a 1D depth-wise convolution, which acts
    as an adaptive low-pass filter as described:
    \begin{equation}
        E_{low}= \text{Conv1d}_{smooth}(\{\tilde{F}_{t}\}_{t=1}^{T}).
    \end{equation}
    This trend signal represents the smooth, continuous evolution of vegetation phenology
    or water level changes. The residual between the original sequence and the
    trend serves as the high-frequency component $E_{high}$, capturing rapid fluctuations
    such as sudden flooding or abrupt environmental changes
    $E_{high}= \tilde{F}- E_{trend}$ The two components are then processed to capture distinct temporal
    features. For the trend component $E_{trend}$, we employ a Gated Recurrent
    Unit (GRU) \citep{chung2014empirical} to model the sequential dependencies and the overall shape
    of the phenological curve, compressing it into a trend embedding $f_{low}$.

    The high-frequency component $E_{high}$ captures sudden reflectance changes
    such as flooding, rainfall events, or human interventions, etc. These signals are sparse and irregular, making them well-suited
    for attention-based modeling. We adopt a multi-head temporal attention
    mechanism, where each attention head specializes in identifying a distinct subset
    of critical transient events. For each attention head $h$, we compute attention
    weights $\alpha_{h}$ as:
    \begin{equation}
        \alpha_{h}=softmax(\frac{q_{h}^{\top}K_{h}}{\sqrt{d_{k}}}),
    \end{equation}
    where $K_{h}$ is the Key matrix projected from $E_{high}$, and $q_{h}$ is a learnable
    query vector. The head-specific event feature is computed as the attention-weighted
    sum:
    \begin{equation}
        o_{h}= \sum_{t}\alpha_{h}[t]V_{h}[t].
    \end{equation}
    The weighted sum of the Value matrix $V_{h}$ yields a feature vector focusing
    on key events. The outputs of all heads are concatenated to form the final
    high-frequency feature:
    \begin{equation}
        f_{high}=Concat(o_{1},o_{2},...,o_{H}).
    \end{equation}
    Finally, the global representation is obtained by concatenating the trend
    and event features
    \begin{equation}
        f_{fused}=Concat(f_{low},f_{high}).
    \end{equation}

    We should note that, unlike the standard SAM, which uses category-agnostic point prompts, our weakly
    supervised setting provides both spatial locations and semantic categories. As shown in Figure \ref{fig:temporal} (b), let $\mathcal{P}_{\text{patch}}=\{(p_{i},l_{i})\}_{i=1}^{N_p}$ denote the
    annotated points within a patch. Each point is encoded into a class-aware token:
    \begin{equation}
        q_{i}=\text{LayerNorm}\big(e^{pt}(p_{i})+e^{cls}(l_{i})\big),
    \end{equation}
    where $e^{pt}(\cdot)$ and $e^{cls}(\cdot)$ denote positional and
    category embeddings, respectively. This yields a set of point tokens $Q=\{q_{i}
    \}_{i=1}^{N_p}$. To handle the variable number of point annotations, we
    aggregate $Q$ into a fixed-length context representation using attention
    pooling. Specifically, a set of $M$ learnable latent queries
    $Z\in\mathbb{R}^{D\times M}$ attends to $Q$ via a cross-attention, producing context tokens $C\in\mathbb{R}^{D\times M}$ that
    summarize the semantic distribution of annotated wetlands within the patch. We
    retain the architecture of the SAM decoder while adapting its interaction
    pattern for dense semantic segmentation. Instead of using prompts to query image
    features, we treat the fused time-aware image features $f_{\text{fused}}$ as
    queries and the context tokens $C$ as keys and values within the decoder:
    \begin{equation}
        F_{\text{dec}}=Decoder_{\text{SAM}}(f_{\text{fused}},C).
    \end{equation}
    This allows each spatial location to attend
    to the global semantic context provided by sparse annotations, resulting in
    a dense feature map $F_{\text{dec}}$ for
    subsequent multi-class prediction.


    \subsection{Spatial branch: iterative boundary refinement}

    To complement the semantic discrimination provided by the temporal branch
    and to enable effective learning of spatial structures under sparse
    supervision, we introduce a spatial branch that explicitly incorporates region-growing–based
    pseudo supervision into the training process as Figure \ref{fig:wetsam} shows. It consists of a learnable
    expanded prediction head that is trained using dense pseudo labels generated
    by a temporal-constraint region growing strategy. This branch leverages spatial
    autocorrelation, where spatially adjacent pixels with similar temporal
    features are likely to belong to the same wetland type, while preventing the
    error propagation commonly observed in static expansion. Given two pixels
    $u$ and $v$ with time-series profiles
    $x_{u}, x_{v}\in \mathbb{R}^{T\times C}$, we measure their temporal similarity
    using cosine similarity:

    \begin{equation}
        \rho(u,v) = \frac{x_{u}\cdot x_{v}}{\lVert x_{u}\rVert_{2}\lVert x_{v}\rVert_{2}}
        .
    \end{equation}
    This metric focuses on the shape of phenological trajectories rather than absolute
    reflectance magnitude. A pixel is assimilated only if it exhibits a highly
    coherent seasonal trajectory with the seed point (i.e., $\rho > \tau$), thereby
    suppressing leakage into spectrally similar but semantically different regions.To
    prevent noise amplification in early training, the first stage performs
    conservative region growing strictly from ground-truth points
    $\mathcal{P}_{gt}$:

    \begin{equation}
        M_{\text{spat}}^{(0)}= \text{Grow}\big(\mathcal{P}_{gt}, \tau\big),
    \end{equation}
    yielding a high-precision initial mask. As the temporal branch becomes more reliable,
    its predictions provide incremental guidance to densify spatial supervision.
    Specifically, at the end of every $K$ training epochs, unlabeled pixels with
    high-confidence temporal predictions are extracted as pseudo-seeds:
    \begin{equation}
        \mathcal{P}_{\text{pseudo}} = \left\{ u \mid \max(f_{\text{temp}}(u)) > 0.95 \right\}.
    \end{equation}

    To further reduce confirmation bias, these seeds are retained only if their
    predicted class is spatially consistent with their $3\times3$ neighborhood. The
    merged seed set $\mathcal{P}_{\text{total}}= \mathcal{P}_{gt}\cup \mathcal{P}
    _{\text{pseudo}}$ enables a refined region growing process:
    \begin{equation}
        M_{\text{spat}}^{(k)}= \text{Grow}\big(\mathcal{P}_{\text{total}}, \tau\big
        ).
    \end{equation}
    The region growing procedure itself is non-differentiable and does not
    participate in backpropagation. It serves solely as a pseudo-label generator
    that expands sparse point annotations into dense supervision maps. These
    expanded labels are then used to supervise a dedicated expanded prediction
    head, enabling the spatial branch to be trained end-to-end despite the non-differentiable
    nature of the region growing process. This expansion is performed locally
    around the newly added pseudo-seeds to ensure efficiency. The updated map $M{_\text{spat}}
    ^{(k)}$ serves as the dense pseudo-label supervision for the subsequent training
    period. Through this iterative feedback loop, the density of reliable seeds increases
    over time, significantly shortening propagation distances and progressively
    sharpening boundaries. Consequently, the spatial branch continually
    suppresses early-stage noise and enhances structural consistency as the
    model converges.

    \subsection{Prediction alignment: consistency regularization}

    Based on the shared decoder feature map, we employ two parallel prediction
    heads corresponding to two complementary supervision pathways: a temporal head
    and a spatial head. Both heads are instantiated as $1\times1$ convolutional
    layers that project the $D$-dimensional decoder features into $K$-class
    logits.

    The \textit{temporal head}, denoted as $f_{\text{temp}}$, is supervised by sparse
    point-level annotations and focuses on learning reliable semantic categorization
    driven by temporal dynamics. It produces a dense probability map
    \begin{equation}
        p_{\text{temp}}= \sigma\big(f_{\text{temp}}(F_{\text{dec}})\big),
    \end{equation}
    where $\sigma$ denotes the softmax function.

    The \textit{spatial head}, denoted as $f_{\text{spat}}$, is supervised by dense
    pseudo labels generated through the iterative region-growing process and focuses
    on capturing spatial continuity and complete object structures. It produces
    \begin{equation}
        p_{\text{spat}}= \sigma\big(f_{\text{spat}}(F_{\text{dec}})\big).
    \end{equation}

    The temporal head is anchored by the high-confidence sparse annotations $\mathcal{P}
    _{gt}$, and is optimized using a point-wise cross-entropy loss:
    \begin{equation}
        \mathcal{L}_{t}= -\frac{1}{|\mathcal{P}_{gt}|}\sum_{(p,l)\in\mathcal{P}_{gt}}
        \log p_{\text{temp}}^{(p,l)}.
    \end{equation}
    In contrast, the spatial head learns from the dense pseudo-labels
    $M_{\text{spat}}^{(k)}$ produced at iteration $k$ of region growing. To mitigate
    class imbalance and boundary noise, we employ the Lov\'asz-Softmax loss:
    \begin{equation}
        \mathcal{L}_{s}= \mathcal{L}_{\text{Lovasz}}\big(p_{\text{spat}}, M_{\text{spat}}
        ^{(k)}\big).
    \end{equation}

    Although the two heads share the same underlying representation, they are driven
    by complementary supervision signals with different reliability and coverage.
    To couple their learning dynamics and to regularize the influence of noisy
    pseudo-labels, we follow \citep{xu2022consistency} to introduce a prediction alignment loss that enforces consistency
    between their dense probability maps:
    \begin{equation}
        \mathcal{L}_{a}= \frac{1}{H\times W}\sum_{i=1}^{H\times W}\big\| p_{\text{temp}}
        ^{(i)}- p_{\text{spat}}^{(i)}\big\|_{2}^{2}.
    \end{equation}

    The overall training objective is given by:
    \begin{equation}
        \mathcal{L}_{total}= \mathcal{L}_{t}+ \mathcal{L}_{s}+ \lambda_{a}
        \mathcal{L}_{a},
    \end{equation}
    where $\lambda_{a}$ controls the strength of the alignment
    regularization.

    \section{Study area and material} \label{sec:dataset}
    \subsection{Study area}
\begin{figure}[h]
        \centering
        \includegraphics[width=\linewidth]{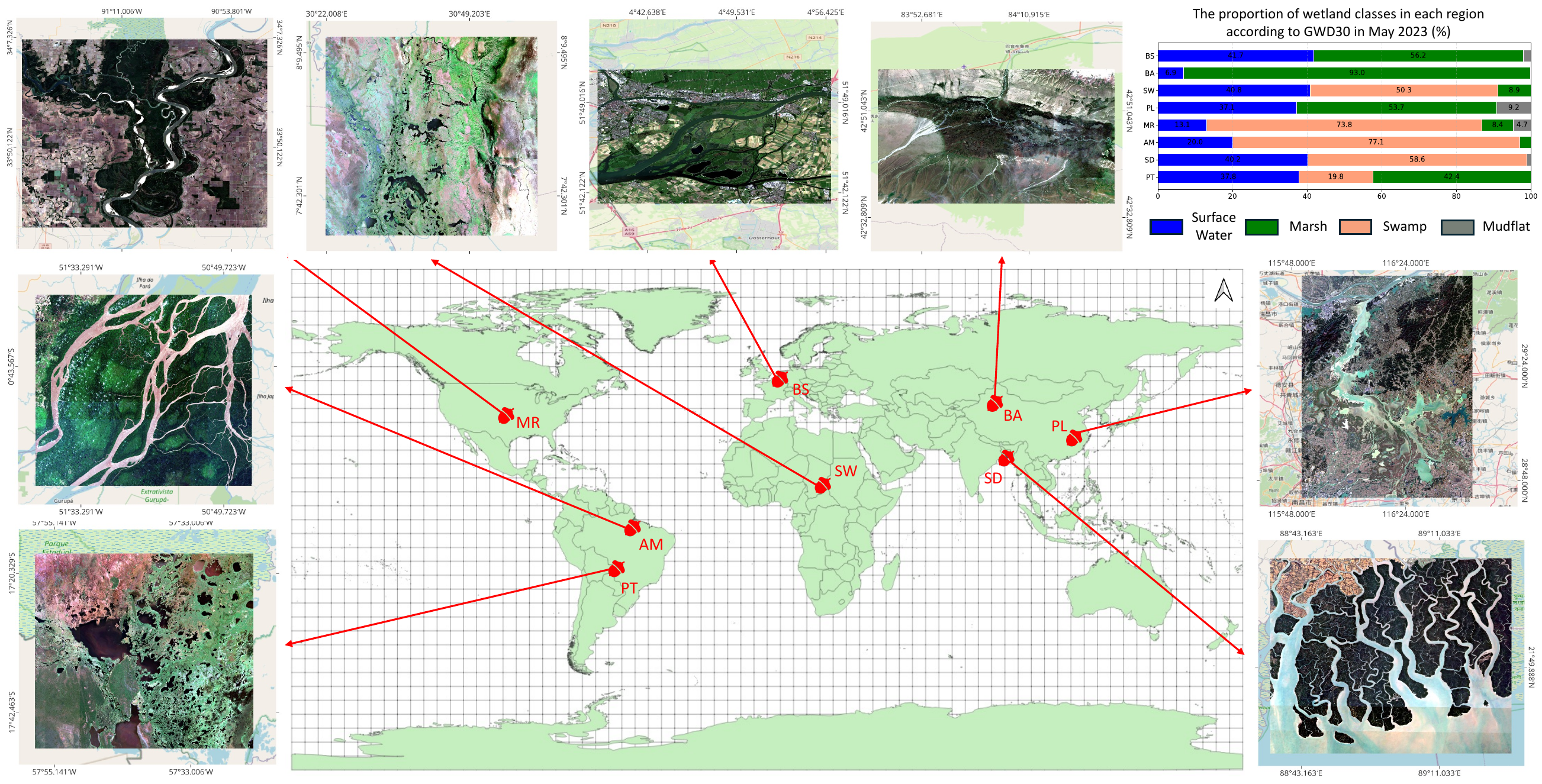}
        \vspace{-1em}
        \caption{The study areas in our study, including 1) Poyang Lake, China (PL);
        2) Mississippi River, United States (MR); 3) Sundarbans, India \& Bangladesh
        (SD); 4) Sudd Wetland, South Sudan (SW); 5) Amazon, Brazil (AM); 6) Biesbosch,
        Netherlands (BS); 7) Pantanal, Brazil (PT); 8) Bayanbulak, China (BA). The
        sub-figure on the top right denotes the distributions of wetland classes
        in each region based on GWD30 \citep{yuan2025gwd30}. }
        \label{fig:study area}
    \end{figure}

    To comprehensively evaluate the effectiveness of the proposed WetSAM
    framework, we conduct experiments across eight selected study areas distributed
    on different continents with different wetland landscapes (Figure
    \ref{fig:study area}): 1) Poyang Lake, China (PL); 2) Mississippi River, United
    States (MR); 3) Sundarbans, India \& Bangladesh (SD); 4) Sudd Wetland, South
    Sudan (SW); 5) Amazon, Brazil (AM); 6) Biesbosch, Netherlands (BS); 7) Pantanal,
    Brazil (PT); 8) Bayanbulak, China (BA).

    These regions are chosen based on the following considerations: First, they encompass
    a broad range of wetland ecosystems across different climatic zones and continents,
    thereby providing a comprehensive and representative testbed for evaluating
    WetSAM. The study areas span tropical humid regions, subtropical monsoon
    regions, temperate maritime zones, arid continental basins, and equatorial savanna
    regions, ensuring exposure to diverse atmospheric and illumination
    conditions. Correspondingly, these regions have multiple kinds of wetlands, including
    mudflats, coastal mangroves, river-fed marshes, and seasonally inundated
    swamps, etc. Such ecological and geographical diversity allows us to
    comprehensively assess the model’s robustness and generalization ability
    across different wetland landscapes. Second, these regions exhibit distinct
    temporal patterns. Some areas experience seasonal inundation cycles (PL, MR,
    PT), while others maintain relatively stable wetland conditions throughout
    the year (SD, BS). These contrasting phenological and hydrological
    trajectories are essential for evaluating the model's capability to capture
    different wetland dynamics. Third, each region covers a large area with
    around 5,000 $km^{2}$, which captures significant spatial heterogeneity and representativeness,
    providing a challenging setting for testing segmentation performance under
    sparse point supervision.

    \subsection{Material}

    A Sentinel-2 Level-2A time-series dataset is used in this study to support
    wetland mapping across the eight selected regions. Sentinel-2 provides multi-spectral
    bands ranging from visible to shortwave infrared with spatial resolutions of
    10–60 m, delivering rich spectral information for monitoring vegetation, water
    bodies, soil moisture, and inland/coastal hydrological processes. All data
    is acquired from the Sentinel-2 Surface Reflectance Harmonized collection on
    Google Earth Engine (GEE) for the period from January 1, 2022 to December 31,
    2023, covering two full hydrological cycles for each region. To ensure high-quality
    observations, we apply a two-stage cloud filtering strategy. First, Sentinel-2
    scenes were filtered using the metadata field, retaining only images with
    cloud cover less than 20\% within each study region. Second, we apply a pixel-level
    cloud and cirrus mask using the QA60 quality assurance band. Cloud and
    cirrus flags are removed via bitwise masking. After cloud masking, three
    bands with 10-m resolution are preserved for analysis: B2 (Blue), B3 (Green),
    B4 (Red). For each study region, a median composite is generated from all available
    cloud-free images to produce 12 representative monthly synthetic images.

\begin{figure}[h]
        \centering
        \includegraphics[width=\linewidth]{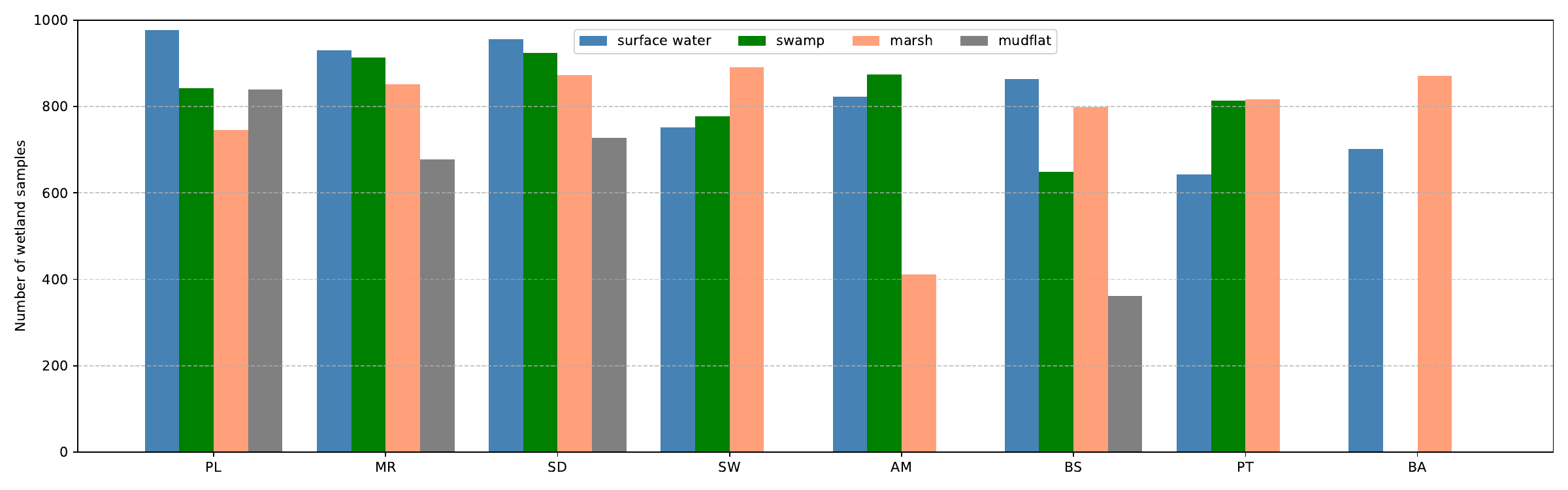}
        \vspace{-1em}
        \caption{The number of wetland samples of each study area.}
        \label{fig:samples}
    \end{figure}

    Additionally, high-resolution Google Earth images and corresponding Sentinel-2
    images are used to annotate and validate the point labels. We focus on four
    major classes, including surface water, swamp, marsh, and mudflat. We manually label approximately 2,000-4,000 points for each study area,
    with balanced representation across classes. Points are uniformly distributed
    within each of the eight study areas to avoid spatial overlap. Figure
    \ref{fig:samples} summarizes the detailed sample information.

    \section{Experiments and evaluation} \label{sec:results}
    \subsection{Experiment settings}
    All models are trained on the weakly supervised setting, where only point-based
    annotations are used as supervision. During training, each time-series image
    is cropped into $256\times256$ patches, and the corresponding sparse points (and
    their region-grown expansions) within each patch serve as the supervision signals
    for the spatial and temporal branches of WetSAM. We adopt an 8:2 split of
    the image patches for training and validation. 

    Training is performed using the PyTorch framework with four NVIDIA GeForce RTX
    4090 GPUs for 50 epochs. The batch size is set to 8. The initial learning
    rate is $1 \times 10^{-5}$ with a weight decay of $4 \times 10^{-5}$, and AdamW
    is used as the optimizer. Standard data augmentation techniques, including random
    flipping and rotation, are applied to enhance data diversity. During training,
    the SAM image encoder is kept frozen to preserve the
    pretrained structural priors of SAM under weak supervision. The hierarchical
    adapters, temporal aggregation module, conditioning injection layers in the decoder,
    as well as both the two segmentation heads are trained end-to-end. For evaluation, independent annotated points are used solely as the test set to quantitatively assess the performance of WetSAM under sparse supervision.

    We use three common metrics to evaluate the performance of WetSAM, i.e.,
    Precision, Recall, and F1-score. The mathematical definition of these metrics
    are:
    \begin{equation}
        Precision = \frac{1}{N}\sum_{n=1}^{N}\frac{TP}{TP+FP}\label{1}
    \end{equation}
    \begin{equation}
        Recall = \frac{1}{N}\sum_{n=1}^{N}\frac{TP}{TP+FN}\label{2}
    \end{equation}
    \begin{equation}
        F1\text{-}score = \frac{1}{N}\sum_{n=1}^{N}\frac{2TP}{2TP+FP+FN}\label{3}
    \end{equation}

    \subsection{Results of WetSAM}
\begin{figure}[h]
        \centering
        \includegraphics[width=\linewidth]{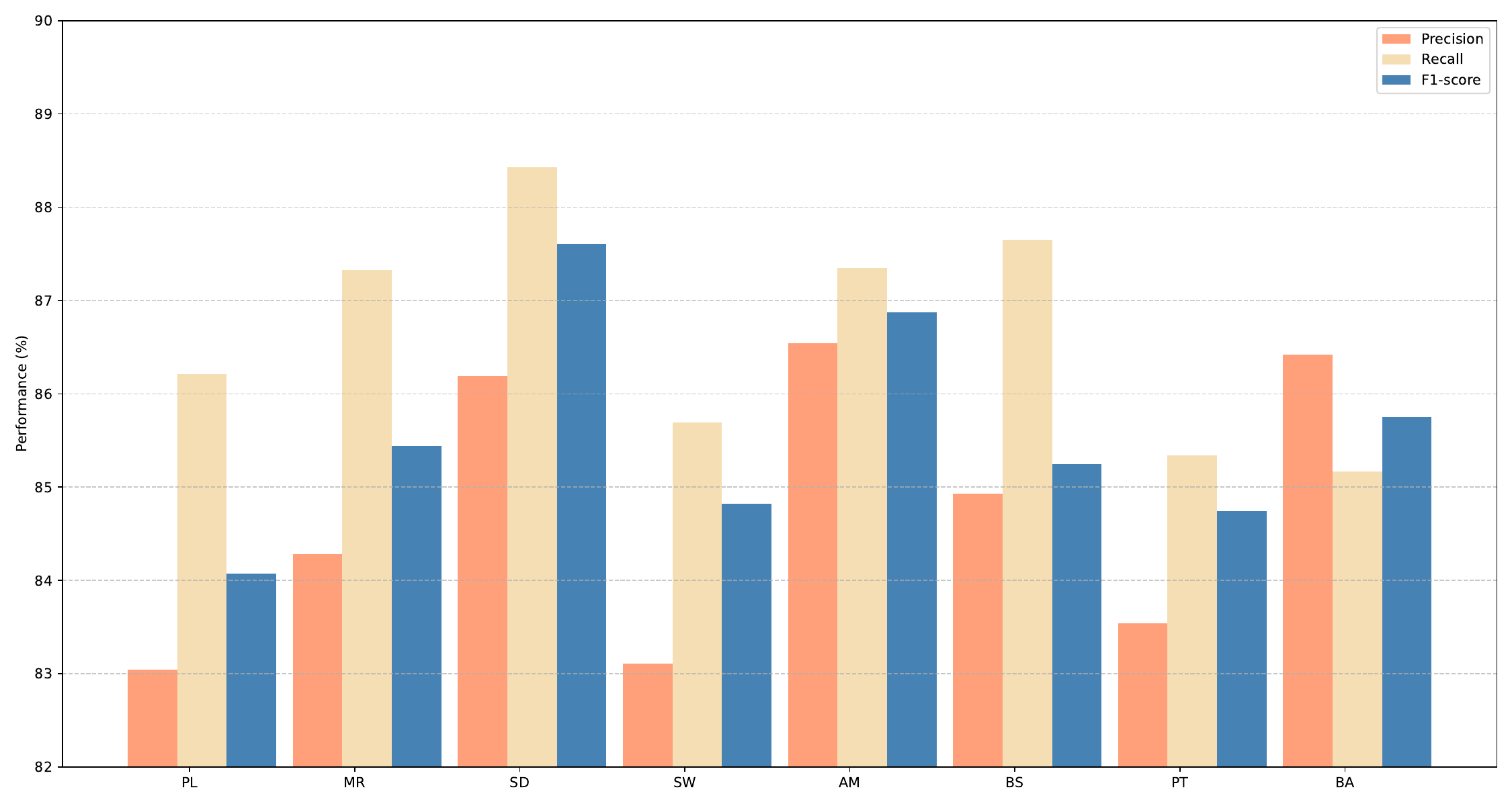}
        \vspace{-1em}
        \caption{The performance of WetSAM in eight study areas, regarding three
        metrics, i.e., Precision, Recall, and F1-score.}
        \label{fig:performance}
    \end{figure}

    We first evaluated the overall segmentation performance of WetSAM across the
    eight study regions. Figure \ref{fig:performance} summarizes the global mean
    Precision, Recall, and F1-score. Under sparse point-level supervision,
    WetSAM achieves an average F1-score of 85.58\%, an average Precision of
    85.26\%, and an average Recall of 86.64\%. The consisent performance in
    eight study areas and three metrics demonstrates that WetSAM is able to
    recover the full spatial extent of wetland patches while maintaining
    accurate boundary delineation, despite being trained solely on sparse annotations. 

    Across individual regions, WetSAM shows stable and robust performance under diverse hydrological and climatic conditions. In stable wetland environments such as Sundarbans (SD) and Biesbosch (BS), WetSAM achieves powerful results with an average F1-score of 86.43\%. In strongly seasonal regions such as Poyang Lake (PL), Sudd wetland (SW), and Pantanal (PT), WetSAM maintains a competitive average F1-score of 84.39\%. Even though there is a slight decrease in these dynamic areas, the performance remains consistently high despite substantial seasonal transitions. This demonstrates WetSAM’s superior ability to resolve temporal spectral ambiguity and handle seasonal hydrological fluctuations by effectively balancing precision and recall. 

    We further report the performance by wetland type in Table \ref{tab:type-level},
    focusing on surface water, marsh, swamp, and mudflat. Surface water achieves
    the highest accuracy among all classes, reflecting its distinctive spectral response
    and relatively stable seasonal behavior. Marsh, swamp and mudflat show slightly
    lower but still robust performance, which is affected by their mixed
    vegetation–water composition and more complex phenological dynamics. Regional
    variability also corresponds well with known ecological differences. For
    example, SD and BS, where wetland conditions remain relatively stable
    throughout the year, exhibit uniformly high accuracy across surface water,
    marsh, swamp, and mudflat. In contrast, highly seasonal systems such as PL
    and PT show mildly reduced accuracy in vegetated and transitional wetland
    types, reflecting the stronger intra-annual variability and more complex boundary
    dynamics in these regions. BA, which lacks swamp and mudflat classes,
    maintains strong performance in surface water and marsh due to its
    relatively homogeneous landscape structure.

    \begin{table*}
        [h]
        \centering
        \caption{Type-level mapping performance (Precision / Recall / F1-score, \%) of
        WetSAM across all eight study regions.}
        \label{tab:type-level} \resizebox{0.85\textwidth}{!}{
        \begin{tabular}{ccccc}
            \hline
            Region  & \textbf{Surface Water} & \textbf{Marsh}        & \textbf{Swamp}        & \textbf{Mudflat}      \\
            \hline
            PL      & 86.24 / 89.41 / 87.80  & 83.84 / 87.01 / 85.40 & 81.84 / 85.01 / 83.39 & 80.24 / 83.41 / 81.79 \\
            MR      & 87.48 / 90.53 / 88.98  & 85.08 / 88.13 / 86.58 & 83.08 / 86.13 / 84.58 & 81.48 / 84.53 / 82.98 \\
            SD      & 89.39 / 91.63 / 90.50  & 86.99 / 89.23 / 88.10 & 84.99 / 87.23 / 86.10 & 83.39 / 85.63 / 84.50 \\
            SW      & 85.61 / 88.19 / 86.88  & 83.11 / 85.69 / 84.38 & 80.61 / 83.19 / 81.88 & --                    \\
            AM      & 89.04 / 89.85 / 89.44  & 86.54 / 87.35 / 86.94 & 84.04 / 84.85 / 84.44 & --                    \\
            BS      & 88.13 / 90.85 / 89.47  & 85.73 / 88.45 / 87.07 & 83.73 / 86.45 / 85.07 & 82.13 / 84.85 / 83.47 \\
            PT      & 86.04 / 87.84 / 86.93  & 83.54 / 85.34 / 84.43 & 81.04 / 82.84 / 81.93 & --                    \\
            BA      & 87.92 / 86.67 / 87.29  & 84.92 / 83.67 / 84.29 & --                    & --                    \\
            \hline
            Average & 87.48 / 89.37 / 88.41  & 84.97 / 86.86 / 85.90 & 82.76 / 85.10 / 83.91 & 81.81 / 84.60 / 83.19 \\
            \hline
        \end{tabular}}
\end{table*}
    
\begin{figure}[h]
        \centering
        \includegraphics[width=\linewidth]{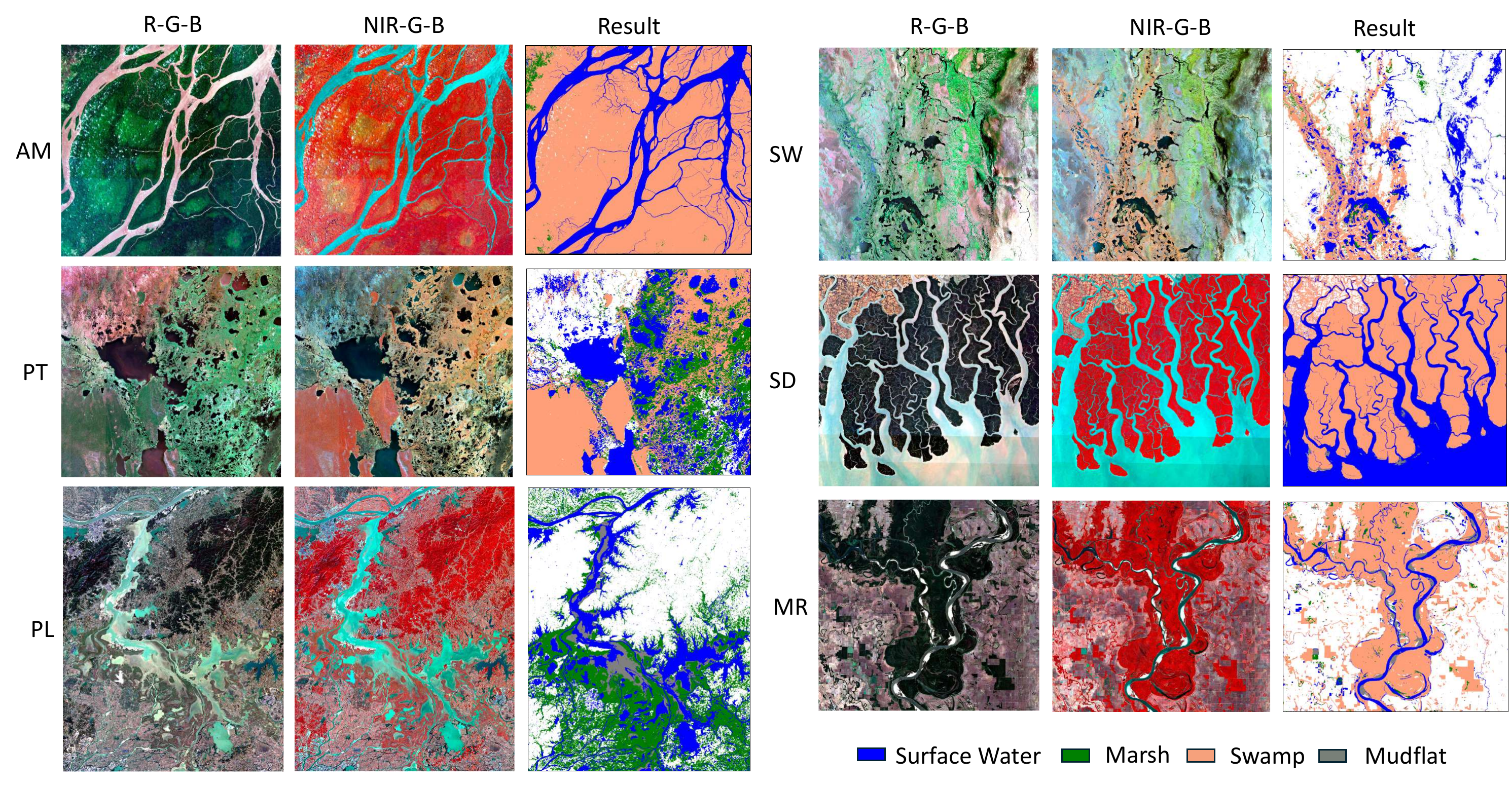}
        \vspace{-1em}
        \caption{The visualization results of WetSAM.}
        \label{fig:result_visual}
    \end{figure}

    Figure \ref{fig:result_visual} presents representative qualitative results
    produced by WetSAM across the study regions. The maps exhibit smooth and coherent
    wetland boundaries, including in fragmented transition zones (e.g., SW, PL) where
    spatial structures are typically complex. Fine-scale wetland features, including
    narrow water channels (e.g., AM, MR), small marsh areas (e.g., PT), and
    spatially sparse wetland patches (e.g., SW, PL), are clearly preserved, indicating
    strong sensitivity to detailed spatial patterns. Overall, the visualization results
    align with the quantitative results and demonstrate that WetSAM can generate
    spatially continuous, structurally meaningful, and temporally stable segmentation
    outcomes under sparse supervision.

    \subsection{Comparative studies}
    We benchmark WetSAM against a comprehensive set of baseline methods, and a
    brief introduction is as follows. The original SAM \citep{kirillov2023segment} serves as prompt-driven
    foundation models capable of zero-shot segmentation but without temporal
    modeling. The SAM 2 \citep{ravi2024sam} supports temporal inputs with continuous visual observations. DINO-SAM \citep{zamboni2025we} enhances SAM with self-supervised DINO features
    to improve mask quality in complex scenes. PSPNet \citep{zhao2017pyramid} represents
    classical semantic segmentation with pyramid pooling for multi-scale context
    aggregation, while SegNext \citep{guo2022segnext} introduces a lightweight yet effective
    design for high-resolution segmentation tasks. CRGNet \citep{xu2022consistency} is a weakly-supervised
    segmentation method using sparse point annotations. UTAE \citep{garnot2021panoptic} is a
    temporal encoder–decoder model tailored for satellite image time series, capable
    of learning temporal dynamics but requiring dense supervision. Together, these
    baselines provide a diverse and representative comparison for evaluating WetSAM
    under sparse annotation settings.

    All baseline results are summarized in Table \ref{tab:comparison}. It should be noted that while most methods are trained directly with sparse point labels, the traditional semantic segmentation models (PSPNet, SegNext, and UTAE) are supervised by pseudo pixel labels expanded from the sparse points using the region-growing algorithm.

Overall, WetSAM achieves the highest F1-score among all methods, reaching an average of 85.58\%. Across individual regions, WetSAM consistently yields strong and robust performance, maintaining F1-scores between 84\% and 88\%. This includes stable wetland environments such as SD and BS, as well as more dynamic systems like PL and SW. In highly seasonal regions (PL, SW, and PT), where substantial intra-annual flooding and vegetation changes typically increase segmentation difficulty, WetSAM maintains a clear advantage over all other baselines.

\begin{table*}
        [t]
        \centering
        \caption{Comparison of F1-score (\%) across eight study regions for all
        baseline methods and WetSAM. }
        \label{tab:comparison} \resizebox{\textwidth}{!}{
        \begin{tabular}{ccccccccccc}
            \hline
            \textbf{Method}   & \textbf{Type}        & \textbf{PL} & \textbf{MR} & \textbf{SD} & \textbf{SW} & \textbf{AM} & \textbf{BS} & \textbf{PT} & \textbf{BA} & \textbf{Average} \\
            \hline
            SAM \citep{kirillov2023segment}      & Foundation model     & 78.14       & 80.21       & 82.55       & 76.12       & 79.34       & 80.01       & 75.22       & 77.89       & 78.69            \\
            SAM 2 \citep{ravi2024sam}    & Foundation model     & 81.54       & 83.02       & 85.12       & 82.11       & 84.45       & 83.05       & 82.54       & 83.37       & 83.15            \\
            DINO-SAM \citep{zamboni2025we} & Foundation model     & 80.26       & 82.35       & 84.41       & 78.36       & 81.55       & 82.47       & 76.98       & 79.54       & 80.74            \\
            PSPNet \citep{zhao2017pyramid}   & Traditional DL model & 72.31       & 75.44       & 77.12       & 70.08       & 73.66       & 74.52       & 68.32       & 72.11       & 73.21            \\
            SegNext \citep{guo2022segnext}  & Traditional DL model & 73.12       & 76.11       & 78.23       & 71.21       & 74.22       & 75.91       & 69.11       & 73.02       & 74.37            \\
            CRGNet \citep{xu2022consistency}   & Weakly supervised    & 77.01       & 79.12       & 81.33       & 75.05       & 78.14       & 79.22       & 73.45       & 76.33       & 77.46            \\
            UTAE \citep{garnot2021panoptic}     & Temporal model       & 79.34       & 81.45       & 83.22       & 77.11       & 80.56       & 81.12       & 76.12       & 78.41       & 79.67            \\
            \hline
            WetSAM     &  Ours   & 84.07       & 85.44       & 87.61       & 84.82       & 86.87       & 85.25       & 84.74       & 85.75       & 85.58            \\
            \hline
        \end{tabular}}
    \end{table*}

The foundation models (SAM, DINO-SAM, and SAM 2) demonstrate promising results, but their performance remains limited in heterogeneous wetland environments. SAM 2, which incorporates a memory attention mechanism for temporal propagation, achieves a significant improvement over the original SAM with an average F1-score of 83.15\%. However, it still falls short of WetSAM, as its general-purpose memory mechanism lacks explicit modeling for domain-specific phenological trends. Traditional models like PSPNet and SegNext show lower and less stable performance, as they are heavily sensitive to the noise and boundary errors present in the pseudo-pixel labels. The temporal model UTAE performs relatively well in dynamic environments but is limited by the quality of the pseudo-supervision, resulting in moderate accuracy.

\begin{figure}[t]
        \centering
        \includegraphics[width=\linewidth]{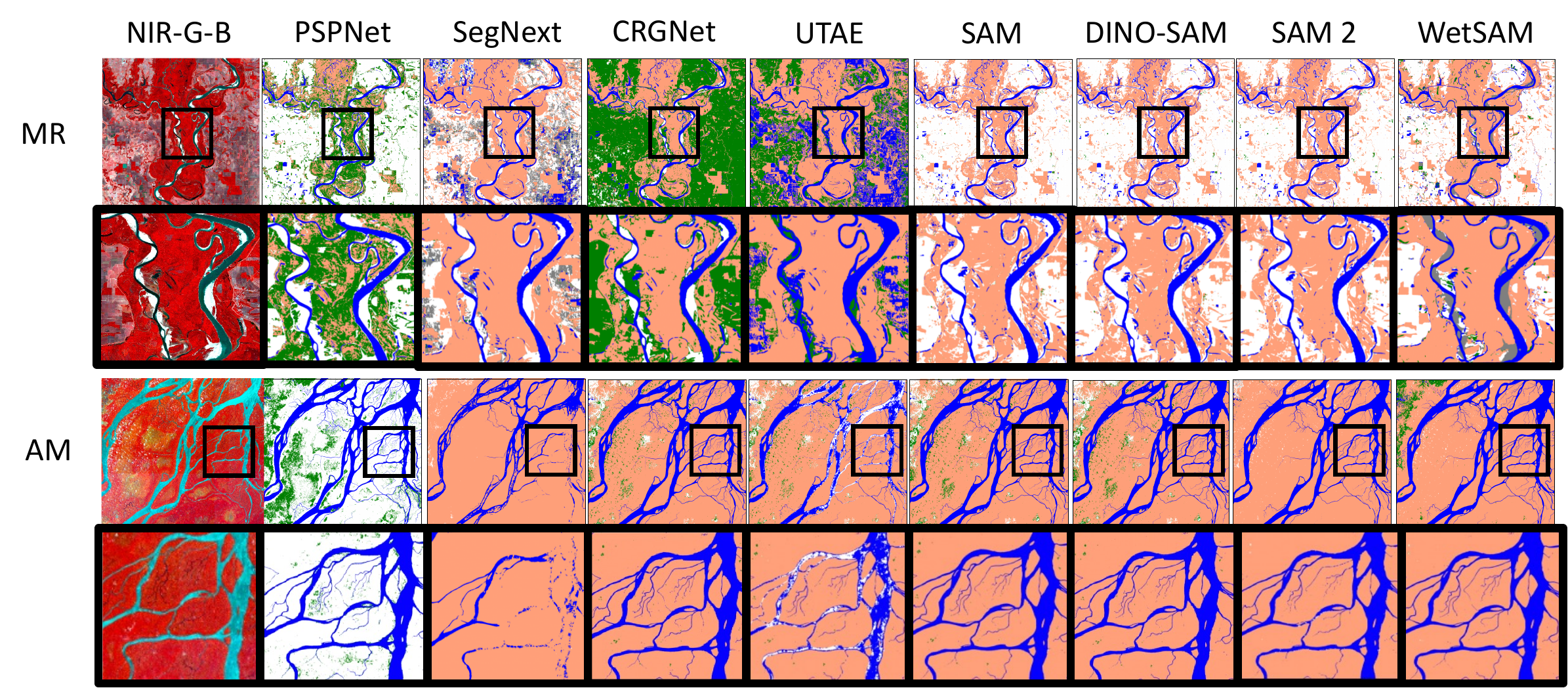}
        \vspace{-1em}
        \caption{The visualization results of comparative studies.}
        \label{fig:result_1}
    \end{figure}

\begin{figure}[t]
        \centering
        \includegraphics[width=\linewidth]{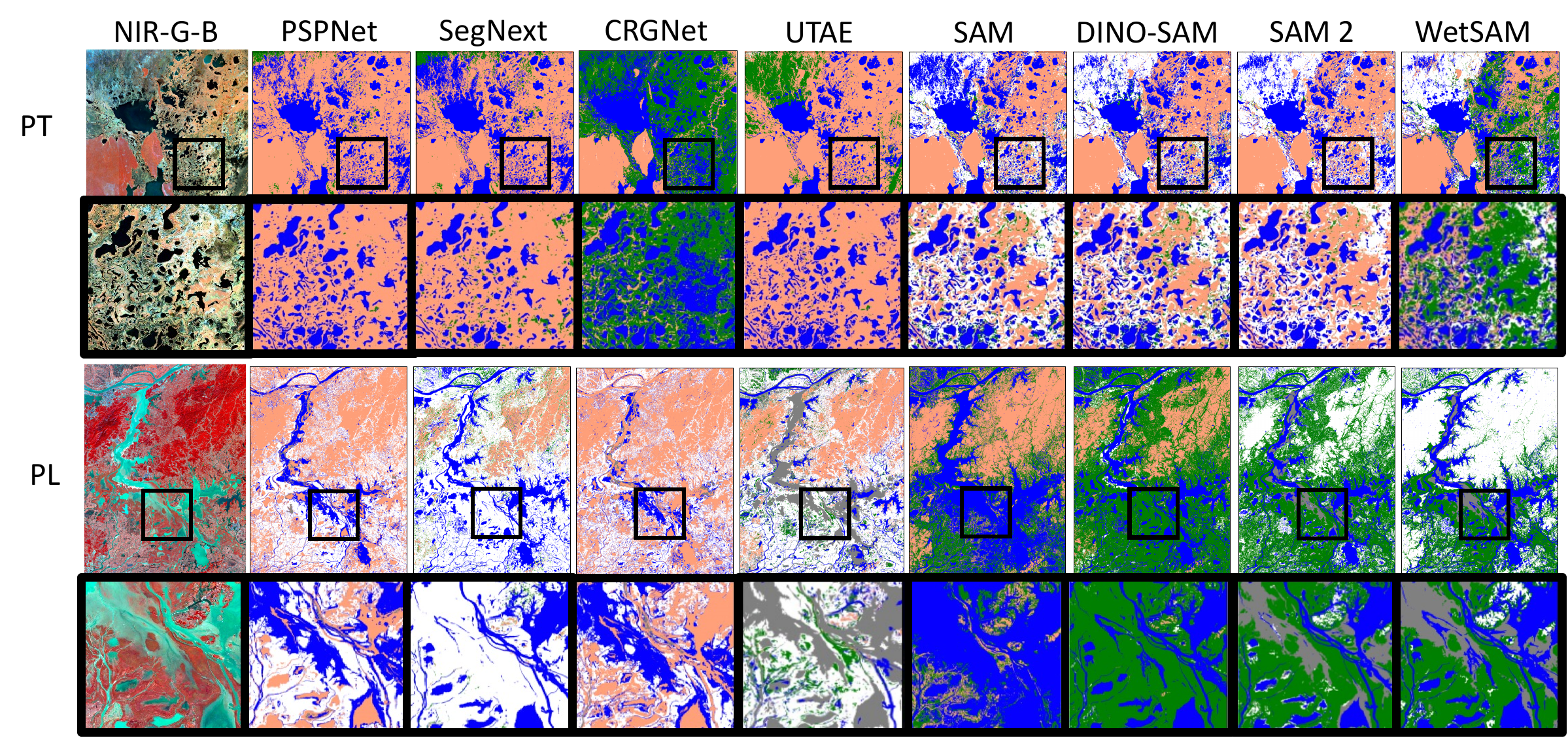}
        \vspace{-1em}
        \caption{The visualization results of comparative studies.}
        \label{fig:result_2}
    \end{figure}

\begin{figure}[t]
        \centering
        \includegraphics[width=\linewidth]{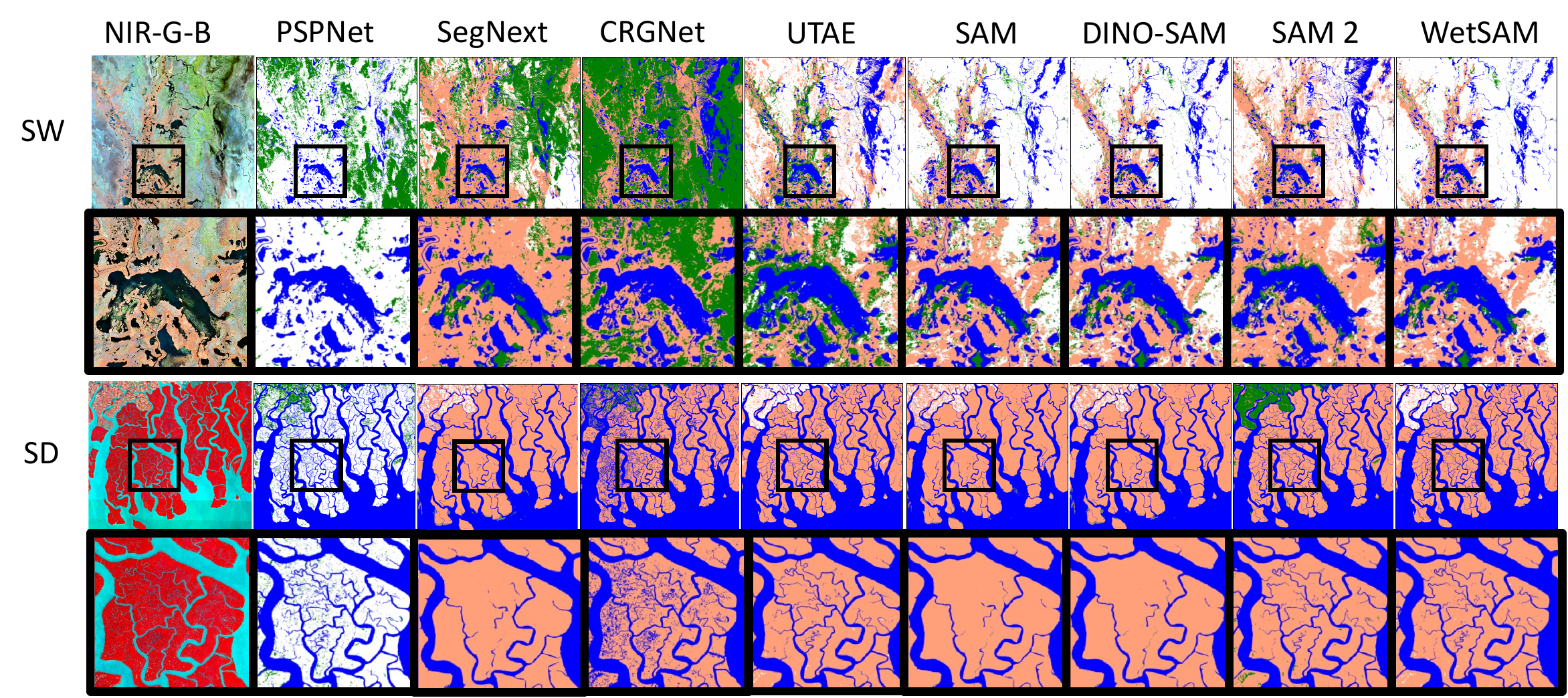}
        \vspace{-1em}
        \caption{The visualization results of comparative studies.}
        \label{fig:result_3}
    \end{figure}

Taken together, these results demonstrate that WetSAM achieves consistently high and robust segmentation accuracy across diverse wetland ecosystems. The performance gap between WetSAM and strong baselines like SAM 2 and UTAE highlights the benefits of jointly temporal dynamics modeling and spatial continuity under sparse annotation settings. Figure \ref{fig:result_1}, Figure \ref{fig:result_2}, and Figure \ref{fig:result_3} provide the visualization results of these comparative studies.


    \section{Discussion} \label{sec:ablation}

    \subsection{Effectiveness of modules in WetSAM}
    To further analyze the contributions of each component in WetSAM, we perform a series of ablation experiments to validate the effectiveness of the proposed modules for accurate wetland mapping under sparse supervision. Below, we detail the impacts of removing each module, and the results are shown in Table \ref{tab:ablation}.
    
    \textbf{Rational for the temporal branch.} Removing the temporal branch results in a model relying solely on spatial features from single-date imagery. In this configuration, the model fails to capture seasonal hydrological cycles essential for discriminating complex wetland types. As shown in Table \ref{tab:ablation}, this leads to the significant performance degradation across all regions, with the average F1-score dropping from 85.58\% to 79.51\%. The largest decreases occurred in highly dynamic regions such as PL, SW, and PT, confirming that temporal context is the indispensable cornerstone for resolving spectral ambiguities caused by seasonal variability.
    
    \textbf{Rational for the spatial branch.} Removing the spatial branch and the region-growing mechanism means the model loses the ability to propagate sparse point annotations into coherent spatial structures. Without dense pseudo-labels for supervision, the model suffers a drop in F1-score to 83.87\%. The segmentation tends to degenerate into scattered predictions around annotated points. However, as observed in Table \ref{tab:ablation}, the performance drop is less severe than that of removing the temporal branch. This suggests that while SAM’s internal spatial priors can aggregate some similar features, the explicit spatial branch is still crucial for reconstructing complete wetland shapes from sparse inputs.
    
    \textbf{Rational for the prediction alignment.} To evaluate the necessity of enforcing consistency between branches, we remove the prediction alignment loss and trained both branches independently. Quantitatively, the average F1-score fell to 84.61\%. Without this mutual regularization, the spatial branch became sensitive to region-growing noise, while the temporal branch failed to benefit from the enriched spatial structures. This highlights the importance of prediction alignment as a bidirectional knowledge distillation mechanism that harmonizes the semantic precision of the temporal branch with the structural completeness of the spatial branch.

\begin{table*}
        [t]
        \centering
        \caption{Ablation study of WetSAM across eight study regions (F1-score, \%).}
        \label{tab:ablation} \resizebox{0.9\textwidth}{!}{
        \begin{tabular}{ccccccc}
            \hline
            \textbf{Region}  & \textbf{w/o Temporal} & \textbf{w/o Spatial} & \textbf{w/o Align} & \textbf{w/o Adapter} & \textbf{w/o Time Emb.} & \textbf{Full Model} \\
            \hline
            PL               & 76.42                 & 82.51                & 83.15              & 81.36                & 83.45                  & 84.07               \\
            MR               & 76.88                 & 83.67                & 84.42              & 82.54                & 84.82                  & 85.44               \\
            SD               & 85.31                 & 85.92                & 86.51              & 84.72                & 86.94                  & 87.61               \\
            SW               & 77.41                 & 83.12                & 83.85              & 82.01                & 84.11                  & 84.82               \\
            AM               & 84.15                 & 85.14                & 85.92              & 84.12                & 86.05                  & 86.87               \\
            BS               & 80.12                 & 83.65                & 84.38              & 82.47                & 84.51                  & 85.25               \\
            PT               & 74.16                 & 82.94                & 83.75              & 81.85                & 83.92                  & 84.74               \\
            BA               & 81.63                 & 84.02                & 84.88              & 82.96                & 85.08                  & 85.75               \\
            \hline
            Average & 79.51                 & 83.87                & 84.61              & 82.75                & 84.86                  & 85.58               \\
            \hline
        \end{tabular}}
\end{table*}
    
    \textbf{Rational for multi-scale adapter.} Removing the hierarchical multi-scale adapter forces the model to use the original SAM encoder without spectral adaptation. Since the original SAM is designed for natural images, it struggles with the satellite imagery and wetland scenarios. As reflected in Table \ref{tab:ablation}, there is a performance degradation of 2.83\% in average F1-score, which underscores the necessity of the multi-scale adapter in bridging the special knowledge gap and recalibrating features to distinguish different wetlands.
    
    \textbf{Rational for time embedding.} To assess the role of temporal positional encoding, we remove the time embedding and relied solely on content-based attention. Without explicit time encoding, the model treats the time series as an unordered set of snapshots rather than an ordered sequence. Quantitatively, we observed a drop in F1-score to 84.86\%, particularly in regions where phenological timing is crucial. This confirms that time embedding effectively guides the model to learn ordered hydrological patterns rather than relying on unordered spectral similarity alone.

    \subsection{Cross-region generalization}

    We further assess the generalization capability of WetSAM through a cross-region evaluation, in which the model is trained on seven regions and directly tested on one unseen regions without any fine-tuning. This experiment simulates real large-scale wetland mapping scenarios where annotated samples may not be available for all areas of interest. 

    As summarized in Table \ref{tab:cross}, WetSAM demonstrates superior robustness across all eight unseen test regions, achieving a stable average F1-score of 82.99\%. Compared to the fully supervised setting, WetSAM exhibits a marginal average performance drop of 2.59\%. In contrast, the traditional temporal model UTAE shows a significant performance collapse when applied to unseen regions with an average of 71.77\%, indicating severe overfitting to site-specific features.

Notably, WetSAM outperforms the zero-shot capabilities of SAM 2 by 2.71\% in average F1-score, particularly in ecologically complex regions like PT and PL. This consistent advantage confirms that WetSAM captures a more universal phenological trends that are shared across geographically distant wetland ecosystems. These results highlight WetSAM's potential as a reliable tool for zero-shot large-scale wetland monitoring in data-scarce regions.

\begin{table*}[h]
    \centering
    \caption{Cross-region generalization performance (F1-score, \%). Each column represents the result when the corresponding region is used as the unseen test set, with the model trained on the remaining seven regions.}
    \label{tab:cross}
    \resizebox{0.7\textwidth}{!}{
    \begin{tabular}{cccccccccc}
        \hline
        \textbf{Method} & \textbf{PL} & \textbf{MR} & \textbf{SD} & \textbf{SW} & \textbf{AM} & \textbf{BS} & \textbf{PT} & \textbf{BA} & \textbf{Average} \\
        \hline
        SAM       & 78.14       & 80.21       & 82.55       & 76.12       & 79.34       & 80.01       & 75.22       & 77.89       & 78.69            \\
        SAM 2     & 79.85       & 81.33       & 83.42       & 78.43       & 81.12       & 81.05       & 77.12       & 79.94       & 80.28            \\
        UTAE      & 71.24       & 73.56       & 75.12       & 69.84       & 72.45       & 73.18       & 67.54       & 71.22       & 71.77            \\
        \hline
        \textbf{WetSAM (Ours)} & \textbf{81.82} & \textbf{83.14} & \textbf{85.41} & \textbf{81.25} & \textbf{84.52} & \textbf{83.37} & \textbf{80.84} & \textbf{83.56} & \textbf{82.99}   \\
        \hline
    \end{tabular}}
\end{table*}

    \subsection{Sensitivity to temporal length}

    To investigate the influence of the input temporal length, we vary the number of time-series images fed into the temporal branch by adjusting the window length $n$ from 1 to 12. Table \ref{tab:time_number} summarizes the results. In relatively stable regions such as SD and BS, the F1-score improves steadily with increasing $n$, as additional observations reinforce spectral stability and reduce noise from clouds or anomalous pixels. For highly seasonal regions such as PL and SW, the performance gain is even more pronounced when $n$ increases from 1 to 6, averaging a 5.12\% gain in F1-score. This is because a larger $n$ allows the temporal aggregation module to more accurately model the long-term phenological trends and abrupt hydrological events. However, we observe that the marginal gains diminish beyond $n=6$, with the improvement from $n=9$ to $n=12$ being less than 1\%. This indicates that excessive temporal redundancy yields limited additional information once the primary seasonal cycles are sufficiently captured. Considering the trade-off between computational efficiency and segmentation accuracy, $n=6$ to $12$ appears to be an optimal range for large-scale wetland mapping using WetSAM.

    \begin{table}[h]
    \centering
    \caption{Performance comparison (F1-score, \%) under different length of input time-series images ($n$) across all eight regions.}
    \label{tab:time_number}
    \resizebox{0.45\textwidth}{!}{
    \begin{tabular}{cccccc}
        \hline
        \textbf{Region} & \textbf{$n=1$} & \textbf{$n=3$} & \textbf{$n=6$} & \textbf{$n=9$} & \textbf{$n=12$ (Full)} \\
        \hline
        PL              & 76.42          & 79.31          & 81.64          & 82.92          & 84.07                  \\
        MR              & 76.88          & 79.52          & 82.81          & 84.73          & 85.44                  \\
        SD              & 85.31          & 86.43          & 87.01          & 87.55          & 87.61                  \\
        SW              & 77.41          & 79.89          & 82.43          & 83.11          & 84.82                  \\
        AM              & 84.15          & 84.64          & 86.03          & 86.42          & 86.87                  \\
        BS              & 80.12          & 82.08          & 84.24          & 84.69          & 85.25                  \\
        PT              & 74.16          & 79.03          & 82.17          & 83.28          & 84.74                  \\
        BA              & 81.63          & 82.54          & 84.22          & 85.19          & 85.75                  \\
        \hline
        Average & 79.51 & 81.68 & 83.82 & 84.74 & 85.57         \\
        \hline
    \end{tabular}}
\end{table}

\subsection{Sensitivity to label density}

To evaluate the data efficiency of WetSAM, we investigate its performance under different densities of point-level annotations. Specifically, we randomly sample a subset of the original point labels at ratios of 10\%, 30\%, 50\%, 70\%, 90\%, and 100\% for training. For each ratio, the sampling process ensures that the labels are equally distributed and cover across all wetland types in each study region. As illustrated in Table \ref{tab:point_number}, WetSAM demonstrates strong robustness even under extremely sparse supervision. With only 10\% of the point labels, the model achieves a promising average F1-score of 79.60\%. This performance is already comparable to several baseline methods trained on much denser information, indicating that the prompt-adapted SAM backbone provides a strong structural prior that effectively compensates for the scarcity of supervision.

We observe that the performance gains are most pronounced as the label ratio increases from 10\% to 50\%, with the average F1-score rising from 79.60\% to 83.87\%. Beyond the 70\% threshold, the improvement curve exhibits a distinct saturation effect. Specifically, increasing the label density from 70\% to 100\% yields only a marginal gain of 0.90\% in the average F1-score. This suggests that WetSAM can effectively capture the core phenological and spatial characteristics of wetlands using only a limited fraction of available annotations. This high data efficiency is particularly valuable for large-scale wetland mapping, as it significantly reduces the human labor required for ground-truth collection while maintaining high segmentation accuracy.

\begin{table*}[h]
    \centering
    \caption{Performance comparison (F1-score, \%) under different percentages of point-level annotations across all eight regions.}
    \label{tab:point_number}
    \resizebox{0.45\textwidth}{!}{
    \begin{tabular}{ccccccc}
        \hline
        \textbf{Region} & \textbf{10\%} & \textbf{30\%} & \textbf{50\%} & \textbf{70\%} & \textbf{90\%} & \textbf{100\%} \\
        \hline
        PL              & 76.87          & 81.20          & 83.27          & 83.68          & 83.90          & 84.07          \\
        MR              & 79.21          & 82.64          & 83.72          & 84.51          & 85.34          & 85.44          \\
        SD              & 83.40          & 85.97          & 86.51          & 87.42          & 87.59          & 87.61          \\
        SW              & 78.62          & 81.07          & 82.79          & 83.62          & 84.38          & 84.82          \\
        AM              & 81.67          & 83.51          & 84.98          & 86.07          & 86.65          & 86.87          \\
        BS              & 80.05          & 82.42          & 83.33          & 84.14          & 84.79          & 85.25          \\
        PT              & 77.46          & 80.94          & 82.69          & 83.44          & 84.01          & 84.74          \\
        BA              & 79.51          & 81.93          & 83.66          & 84.51          & 85.23          & 85.75          \\
        \hline
        Average & 79.60 & 82.46 & 83.87 & 84.67 & 85.24 & 85.57 \\
        \hline
    \end{tabular}}
\end{table*}

\subsection{Sensitivity to label noise}

In practice, wetland point annotations are possible to be subject to errors due to the spectral similarities between classes or seasonal shifts. To evaluate the fault tolerance of WetSAM, we introduce random label noise at ratios from 5\% to 60\%. As demonstrated in Table \ref{tab:label_noise}, WetSAM maintains a stable performance when the noise ratio is within 25\%. The average F1-score only experiences a negligible decrease of 1.96\% (from 85.43\% to 83.47\%). However, a sharp drop in performance is observed once the noise exceeds the 30\% threshold, where the erroneous signals begin to overwhelm the true labels. Notably, seasonal regions (e.g., PL, PT, and SW) exhibit much higher sensitivity to extreme noise compared to stable regions. In these dynamic environments, high noise ratios lead to a failure in temporal modeling, causing the F1-score to plunge by over 20\% as the noise moves from 30\% to 50\%. In contrast, stable regions like SD and AM show a more buffered degradation. These results suggest that while WetSAM is highly robust to moderate errors, maintaining label accuracy remains crucial in complex environments.

\begin{table*}[h]
    \centering
    \caption{Robustness analysis of WetSAM under different label noise levels (5\% to 60\%). "Seasonal" regions (PL, MR, SW, PT) exhibit higher sensitivity to noise beyond the 30\% threshold.}
    \label{tab:label_noise}
    \resizebox{0.61\textwidth}{!}{
    \begin{tabular}{c|ccccccccc}
        \hline
        \textbf{Region} & \textbf{5\%} & \textbf{10\%} & \textbf{15\%} & \textbf{20\%} & \textbf{25\%} & \textbf{30\%} & \textbf{40\%} & \textbf{50\%} & \textbf{60\%} \\
        \hline
        PL       & 83.91         & 83.82          & 83.65          & 82.25          & 82.88          & 75.12          & 62.34          & 50.15          & 41.22          \\
        MR       & 85.24         & 85.15          & 84.98          & 83.64          & 84.15          & 77.34          & 64.88          & 53.42          & 44.15          \\
        SD       & 87.55         & 87.48          & 87.31          & 86.11          & 86.82          & 82.45          & 74.12          & 66.89          & 58.42          \\
        SW       & 84.65         & 84.51          & 84.32          & 82.98          & 83.45          & 76.05          & 63.15          & 51.27          & 42.06          \\
        AM       & 86.75         & 86.62          & 86.44          & 85.16          & 85.91          & 81.82          & 73.54          & 65.14          & 57.21          \\
        BS       & 85.11         & 85.02          & 84.88          & 83.57          & 84.22          & 79.76          & 71.22          & 63.45          & 55.33          \\
        PT       & 84.58         & 84.44          & 84.18          & 82.79          & 83.21          & 75.43          & 61.92          & 49.85          & 40.54          \\
        BA       & 85.62         & 85.54          & 85.35          & 84.06          & 84.76          & 80.12          & 72.43          & 64.91          & 56.12          \\
        \hline
        Average & 85.43 & 85.32 & 85.14 & 84.79 & 83.47 & 78.51 & 68.08 & 58.14 & 49.38 \\
        \hline
    \end{tabular}}
\end{table*}

\subsection{Limitations}

Despite the promising performance of WetSAM in weakly supervised wetland mapping, several limitations remain to be addressed in future research. First, the reliance on optical satellite imagery renders the framework susceptible to atmospheric conditions. Although we apply strict cloud filtering, persistent cloud cover in tropical or monsoon-dominated regions can still disrupt the continuity of time series. Integrating All-Weather Synthetic Aperture Radar (SAR) data into the framework to compensate for optical data gaps represents a critical direction for improvement. Second, while WetSAM demonstrates robustness to moderate label noise, our robustness analysis reveals a performance bottleneck under high-noise conditions. This suggests that the correctness of the initial sparse prompts is needed to guide the learning of complex temporal patterns. Third, to leverage the structural priors of the foundation model, we freeze the image encoder of SAM during training. While the hierarchical adapter effectively bridges the domain gap, the frozen backbone may still limit the model's ability to learn features that are fundamentally different from natural images, specifically when extending the framework to other data modalities such as SAR data.

These limitations directly point to several future directions for our future work. First, to mitigate the impact of cloud cover and data gaps in optical imagery, it is necessary to extend the framework to support multi-modal data fusion, specifically by integrating Sentinel-1 SAR data to enable all-weather wetland monitoring. Second, label quality is important; therefore, we aim to construct a reliable global wetland sample database using automatic ways. Finally, we envision exploring parameter-efficient fine-tuning strategies to align the foundation model's representation space with the specific spectral characteristics of multi-modal remote sensing data.

    \section{Conclusion} \label{sec:conclusion}

    In this study, we proposed WetSAM, a novel dual-branch framework designed to address the critical challenge of accurate wetland mapping under extremely sparse point-level supervision. By extending the Segment Anything Model (SAM) with a hierarchical multi-scale adapter and a dynamic temporal aggregation module, WetSAM successfully bridges the gap between static foundation models and dynamic satellite image time series. The proposed framework effectively synergizes temporal and spatial information: the temporal branch leverages phenological trends to resolve spectral ambiguities, while the spatial branch reconstructs coherent boundaries via a temporal-constrained region-growing strategy. A bidirectional prediction alignment further ensures that these two complementary views mutually reinforce each other, resulting in robust segmentation. Extensive experiments across eight geographically and ecologically diverse wetland regions demonstrate the superiority of our approach. WetSAM consistently outperforms state-of-the-art baselines, achieving an average F1-score of 85.58\% and demonstrating remarkable stability even in highly seasonal environments. Crucially, our results highlight the framework's data efficiency; it retains high performance even when trained with only 70\% of the available point annotations, making it a practical solution for large-scale applications where ground truth is scarce. Overall, WetSAM offers a scalable, cost-effective pathway for large-scale wetland monitoring, contributing to better conservation and management of the world's vital wetland resources. In the future, we plan to incorporate multi-modal data fusion, particularly integrating SAR data for all-weather monitoring. We also aim to enhance the model's adaptability by exploring parameter-efficient fine-tuning strategies to unfreeze the image encoder, thereby better aligning the backbone with remote sensing domains.

    \bibliography{mybibfile}
\end{document}